\documentclass{article}

\usepackage[final,nonatbib]{neurips_2020_tda}

\usepackage[utf8]{inputenc} % allow utf-8 input
\usepackage[T1]{fontenc}    % use 8-bit T1 fonts
\usepackage{hyperref}       % hyperlinks
\usepackage{url}            % simple URL typesetting
\usepackage{booktabs}       % professional-quality tables
\usepackage{amsfonts}       % blackboard math symbols
\usepackage{nicefrac}       % compact symbols for 1/2, etc.
\usepackage{microtype}      % microtypography
\usepackage{subfigure}
\usepackage{graphicx}
\usepackage{mathtools}
\usepackage{amsthm}

\title{Regularization of Persistent Homology Gradient Computation}

\author{Padraig Corcoran {\normalfont \&} Bailin Deng\\School of Computer Science \& Informatics\\Cardiff University\\Wales}

\begin{document}

\maketitle

\begin{abstract}
Persistent homology is a method for computing the topological features present in a given data. Recently, there has been much interest in the integration of persistent homology as a computational step in neural networks or deep learning. In order for a given computation to be integrated in such a way, the computation in question must be differentiable. Computing the gradients of persistent homology is an ill-posed inverse problem with infinitely many solutions. Consequently, it is important to perform regularization so that the solution obtained agrees with known priors. In this work we propose a novel method for regularizing persistent homology gradient computation through the addition of a grouping term. This has the effect of helping to ensure gradients are defined with respect to larger entities and not individual points.
\end{abstract}

\section{Introduction}
Persistent homology is a computational method from the field of applied topology for computing the topological features present in a given data~\cite{edelsbrunner2010}. Informally, the features in question relate to the number and scale of connected components and holes of different dimensions in the data. The direct application of persistent homology has proven to be a useful method in the analysis of many different types of data including image~\cite{carlsson2008local}, health~\cite{nicolau2011topology} and network data~\cite{corcoran2020stable}.

Given the recent advances and interest in neural networks or deep learning, there exists a trend of attempting to integrate existing computational methods as computational steps in this framework. This includes, for example, the integration of integer programming~\cite{ferber2020mipaal} and shortest path methods~\cite{berthet2020learning} with deep learning. In these works, deep learning usually acts as a preprocessing step to the method in question where it performs representation learning. Deep learning requires all computational steps to be differentiable such that gradients can be back-propagated through each step and used to update the method parameters. Therefore, for a given computational method to be integrated with deep learning the method in question must be differentiable. Many useful computational methods, such as integer programming, in their native form are not differentiable. Consequently much work has been investing in making such computational methods differentiable so that they may be integrated with deep learning. That is, developing methods for computing the gradients of the input with respect to the output of the methods in question.

As discussed above, persistent homology has proven to be a useful computational method. Therefore integrating it with deep learning has much potential. However, in its native form, persistent homology is not differentiable. Therefore recently there has been much interest in attempting to make it differentiable so that it may be integrated with deep learning~\cite{chen2019topological, wangtopogan, gabrielsson2020topology, Moor19Topological}. Computing gradients of persistent homology outputs with respect to inputs is an inverse problem~\cite{oudot2020inverse, solomon2020fast}. Like many inverse problems, it is ill-posed with infinitely many solutions. This point is obvious when one considers that infinitely many different datasets can have the same number and scale of connected components and holes of different dimensions. When attempting to solve an arbitrary ill-posed inverse problem, if one does not consider the ill-posed nature of the problem, the solution obtained may not agree with known priors. A common approach to overcome this issue is to perform regularization which biases the solution toward known priors. This approach is commonly used to solve inverse problems in the field of image processing. In this field two commonly used regularization approaches are total variation (TV) and total generalized variation (TGV) which lead to piecewise constant and piecewise linear images respectively~\cite{lunz2018adversarial}.

\begin{figure}
\begin{center}
\subfigure[]{\includegraphics[height=3.5cm]{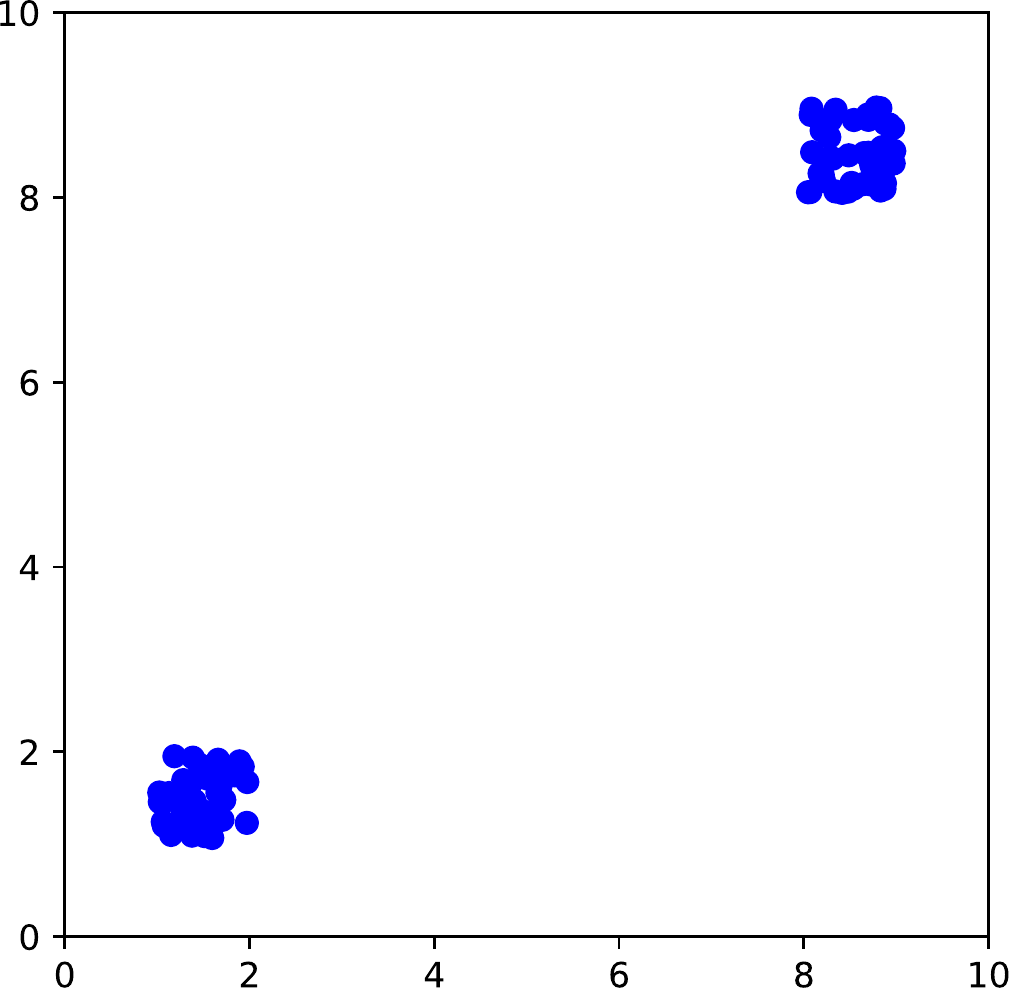}
\label{fig:two_cluster}}
\hspace{1cm}
\subfigure[]{\includegraphics[height=3.5cm]{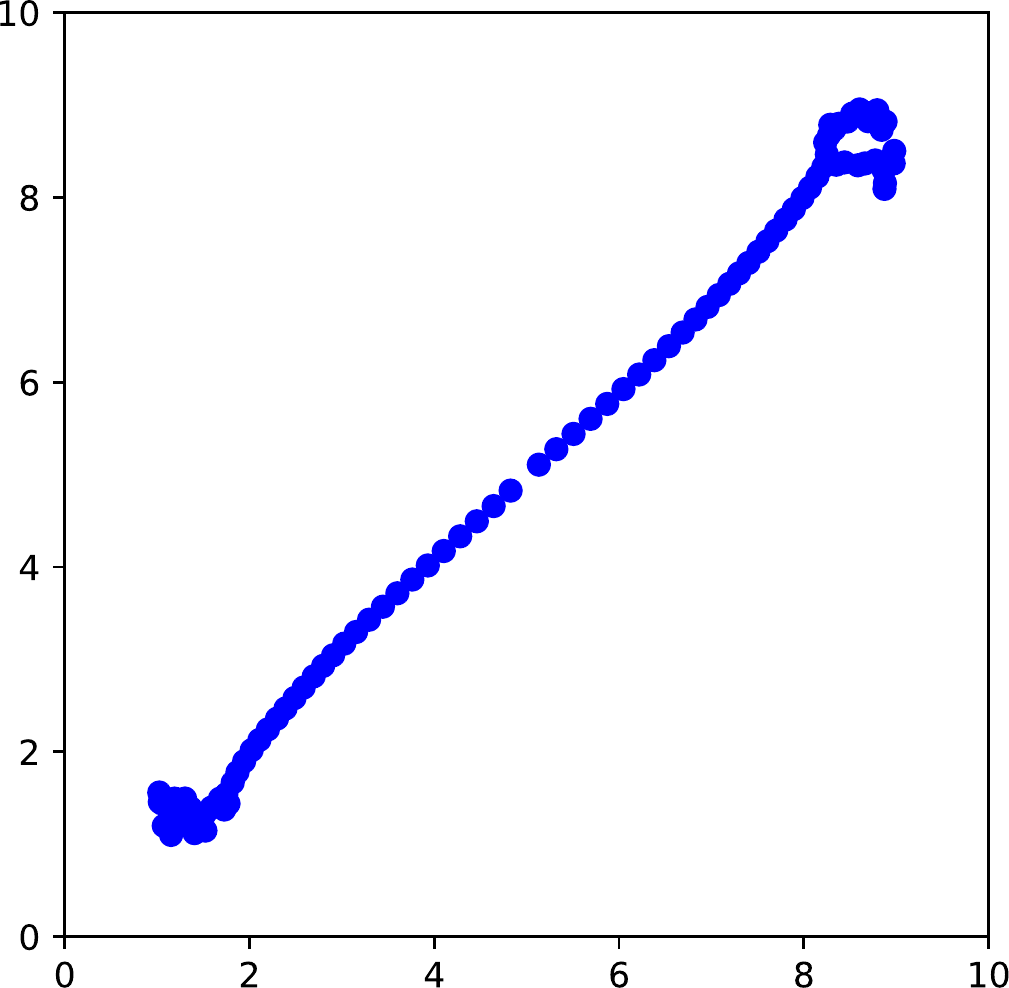}
\label{fig:two_cluster_zero_smooth}}
\hspace{1cm}
\subfigure[]{\includegraphics[height=3.5cm]{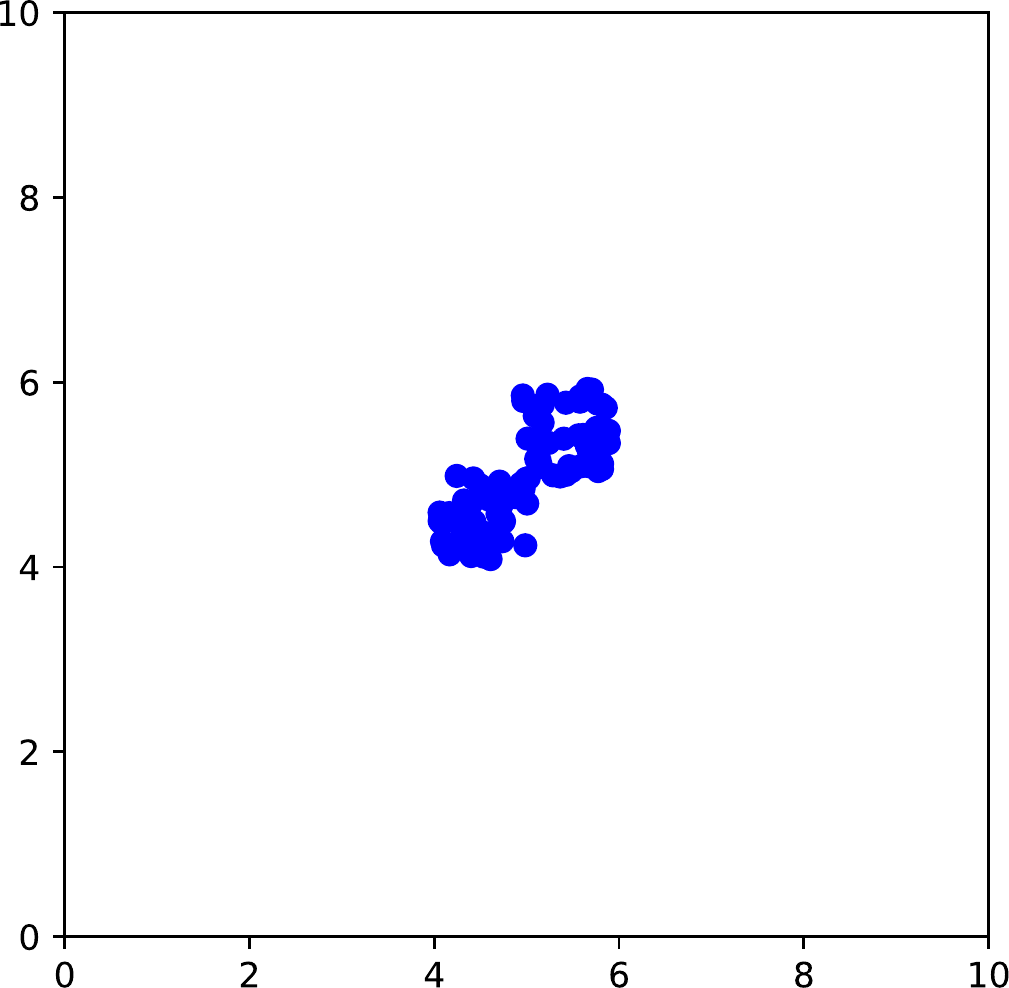}
\label{fig:two_cluster_smooth}}
\\
\subfigure[]{\includegraphics[height=3.5cm]{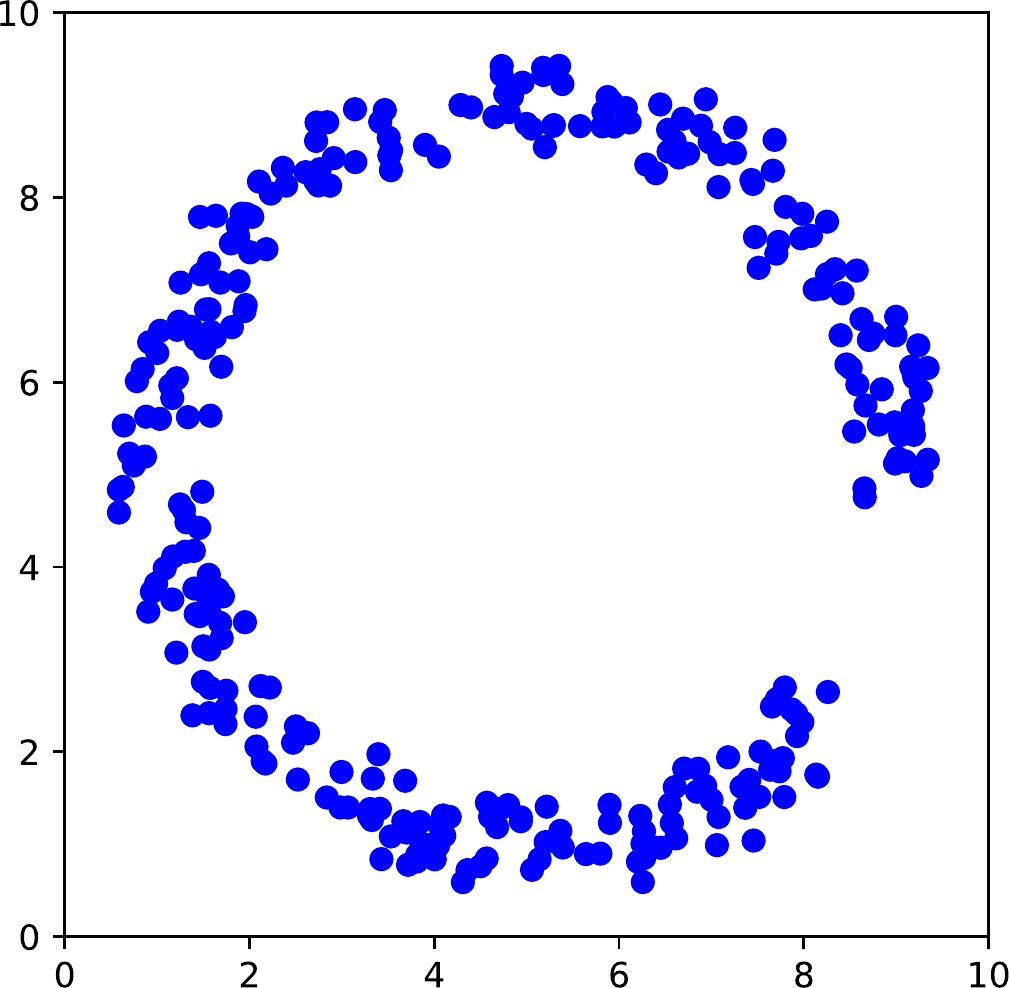}
\label{fig:circle}}
\hspace{1cm}
\subfigure[]{\includegraphics[height=3.5cm]{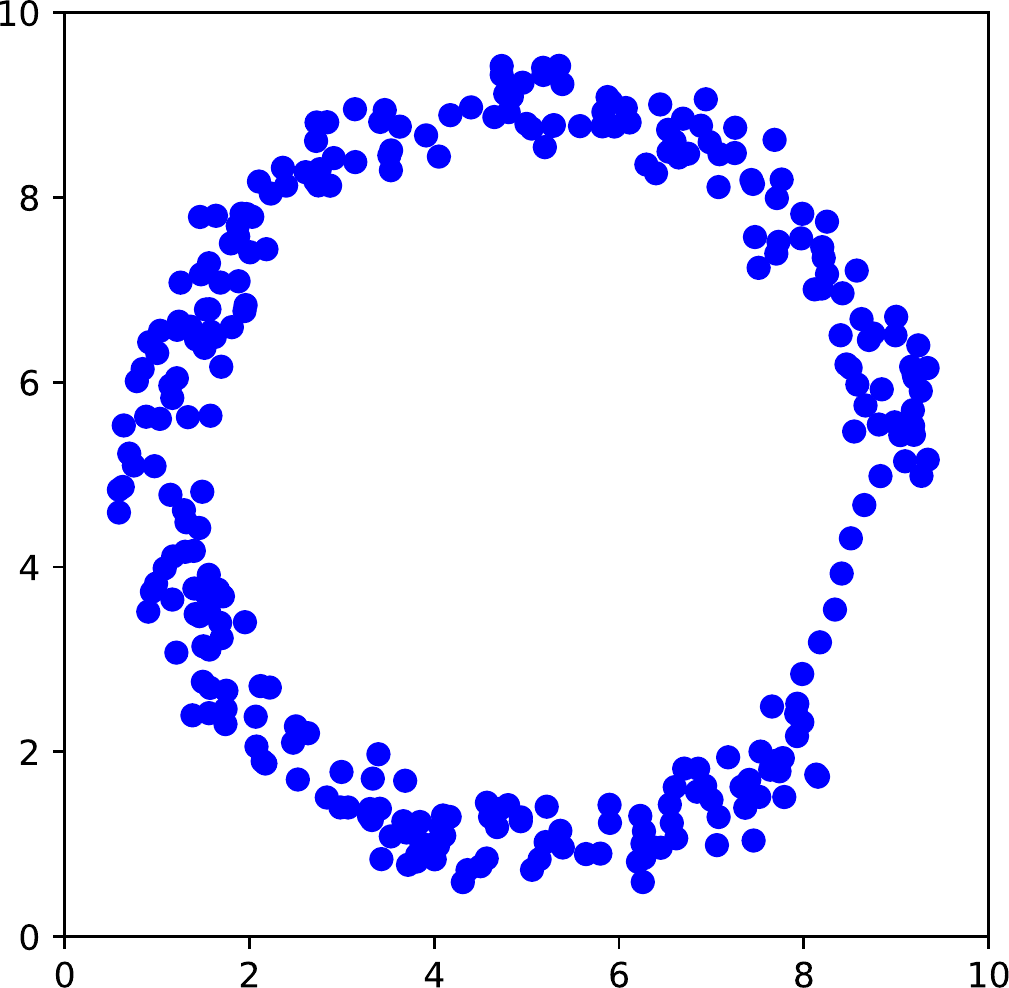}
\label{fig:circle_zero_smooth}}
\hspace{1cm}
\subfigure[]{\includegraphics[height=3.5cm]{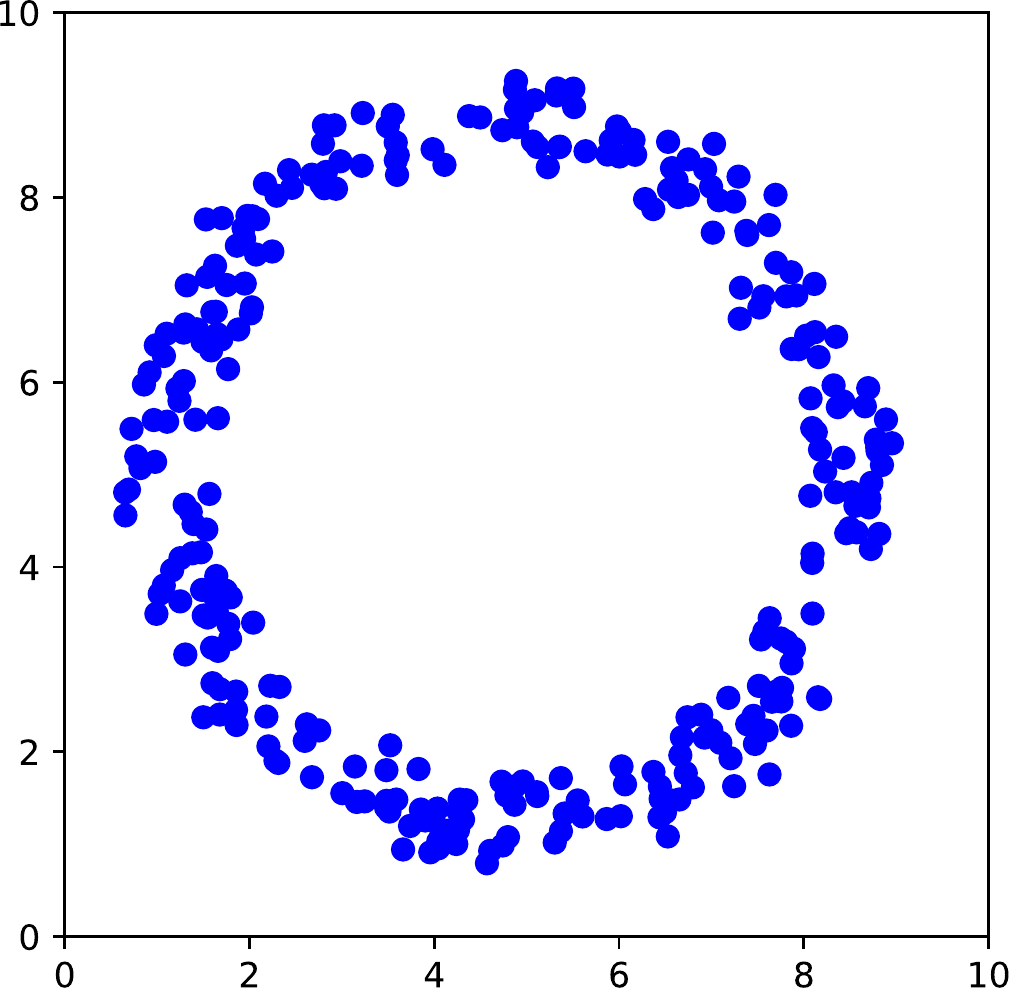}
\label{fig:circle_smooth}}
\caption{Two sets of points in $\mathbb{R}^2$ containing 100 and 300 elements are displayed in (a) and (d) respectively. Each set of points has an associated loss function measuring the distance between the two clusters and two horseshoe ends respectively. The results of minimizing these loss functions without regularization are displayed in (b) and (e) respectively. The results of minimizing these loss functions with the proposed regularization are displayed in (c) and (f) respectively.}
\label{fig:intro_shapes_single_component}
\end{center}
\end{figure}

To the authors knowledge, the use of regularization when computing gradients of persistent homology has yet to be considered. Consequently, the use of current methods for computing such gradients can lead to solutions which do not agree with known priors. For example, consider the two dimensional point dataset in Figure~\ref{fig:two_cluster} which contains two compact clusters. If we compute the gradients of persistent homology using current methods and in turn minimize a loss function measuring the distance between the clusters (see Appendix~\ref{appx:lossfunctions}), we obtain the result in Figure~\ref{fig:two_cluster_zero_smooth}. Although this result approaches the minimization of the loss function in question, it does not agree with a reasonable prior that changes to topology features should be made at the level of larger entities and not individual points. In this case the entities in question are the two clusters. As a second example, consider the two dimensional point dataset in Figure~\ref{fig:circle} which contains a single horseshoe shaped cluster. If we compute the gradients of persistent homology using current methods and in turn minimize a loss function measuring the width of the horseshoe opening (see Appendix~\ref{appx:lossfunctions}), we obtain the result in Figure~\ref{fig:circle_zero_smooth}. Again, although this result approaches the minimization of the loss function in question, it does not agree with a reasonable prior that changes to topology features should be made at the level of larger entities. In this case the entities in question are the two ends of the horseshoe.

The artefacts in the above two example results are a consequence of the fact that persistent homology gradients are defined with respect to individual points and not larger entities. This motivates the following insight. When computing persistent homology gradients, this computation should be regularized through the addition of a grouping term such that gradients are defined with respect to larger entities and not individual points. In this article we propose a novel method for regularizing the computation of persistent homology gradients which achieves this goal. The result of applying this method to the two dimensional datasets in Figures~\ref{fig:two_cluster} and \ref{fig:circle} with the respective loss functions described above are displayed in Figures~\ref{fig:two_cluster_smooth} and \ref{fig:circle_smooth} respectively. It is evident that in the case of these two examples, the proposed method does not exhibit the artefacts encountered above.

The layout of this paper is as follows. In Section~\ref{sec:gradient_compute} we describe the proposed method for regularizing the computation of persistent homology gradients. In Section~\ref{sec:conclusions} we briefly draw some conclusions from this work.

\section{Regularized Gradient Computation}
\label{sec:gradient_compute}
This section is structured as follows. In Section~\ref{sec:gradient_compute:background} we briefly review necessary background material on persistent homology and describe a current method for computing gradients with respect to persistent homology. In Section~\ref{sec:gradient_compute:regularization} we describe the proposed method for performing regularization of this gradient computation.

\subsection{Persistent Homology \& Gradient Computation}
\label{sec:gradient_compute:background}
An (abstract) simplicial complex $\mathcal{K}$ is a finite collection of sets such that for each $\sigma \in \mathcal{K}$ all subsets of $\sigma$ are also contained in $\mathcal{K}$. Each element $\sigma \in \mathcal{K}$ is called a $k$-simplex where $k = \left\vert{\sigma}\right\vert - 1$ is the dimension of the simplex. Given a finite set of points $X = \lbrace x_1, \dots, x_m \rbrace$ in $\mathbb{R}^n$, the corresponding Rips complex $\mathcal{R}_r$ for a specified $r \geq 0$ radius value is defined as follows: 
\begin{equation}
	\label{eq:rips}
	\mathcal{R}_r(X) = \lbrace \sigma \subseteq X : \forall x,y \in \sigma, \| x-y \| \leq r \rbrace
\end{equation}
The $k$-simplices in this simplicial complex equal unordered ($k+1$)-tuples of points which are pairwise within distance $r$~\cite{ghrist2008barcodes}.
Computing the homology of $\mathcal{R}_r(X)$ returns the homology groups $H_i$ for each natural number $i$. An element of $H_i$ represents the existence of an $i$-dimensional hole in $\mathcal{R}_r(X)$. That is, an element of $H_0$ represents the existence of a path-component in $\mathcal{R}_r(X)$ while an element of $H_1$ represents the existence of a one dimensional hole in $\mathcal{R}_r(X)$~\cite{otter2017roadmap}. A Rips filtration of $X$ is a finite sequence of $N$ Rips complexes $\mathcal{R}_{r_1}(X), \dots, \mathcal{R}_{r_N}(X)$ associated with an increasing sequence $r_1, \dots, r_N$ of radius values. A Rips filtration induces a sequence of inclusion maps defined as:
\begin{equation}
\label{eq:filtration}
\mathcal{R}_{r_1}(X) \xhookrightarrow{} \mathcal{R}_{r_2}(X) \xhookrightarrow{} \dots \xhookrightarrow{} \mathcal{R}_{r_N}(X)
\end{equation}
Given a Rips filtration, instead of computing the homology of each Rips complex in the sequence independently, persistent homology computes the homology of the inclusions $\mathcal{R}_p(X) \xhookrightarrow{} \mathcal{R}_{q}(X)$ for all $p<q$~\cite{otter2017roadmap}. The result of this computation is a set of persistence diagrams $\lbrace D_0, \dots, D_h \rbrace$ where $D_i$ corresponds to the homology group $H_i$. Each persistence diagram is a multiset of points $\lbrace (p,q) \in \mathbb{R}^2, p \leq q \rbrace$ where $(p,q) \in D_i$ represents the existence of an element of $H_i$ appearing in $\mathcal{R}_p(X)$ and subsequently disappearing in $\mathcal{R}_q(X)$. The value $q-p$ is called the persistence of the element in question.

Let $(p,q)$ be an element in a given persistence diagram $D_i$. There exists a map $\gamma: \mathbb{R} \rightarrow \mathcal{R}_{r}(X)$ which maps $p$ and $q$ to simplices in the Rips filtration whose addition results in the appearance and disappearance respectively of the corresponding element in $H_i$~\cite{leygonie2019framework, gabrielsson2020topology}. Therefore, one can adjust the values $p$ and $q$ by adjusting the radius values at which the simplices $\gamma(p)$ and $\gamma(q)$ respectively are added in the Rips filtration. Specifically, the radius value at which a simplex is added in the Rips filtration is defined by the following map $\delta: \mathcal{R}_{r} \rightarrow \mathbb{R}$: 
\begin{equation}
	\label{eq:simplex_radius}
	\delta(\sigma) = \max_{x,y \in \sigma} \| x-y \|
\end{equation}
Both the maps $\gamma$ and $\delta$ are differentiable and in turn the map $\gamma \circ \delta$ is differentiable~\cite{leygonie2019framework}. Let $l$ be a real valued differentiable loss function which is a function of $D_i$. The map $l \circ \gamma \circ \delta$ is in turn differentiable and can be minimized using any gradient based optimization technique.

\subsection{Regularization}
\label{sec:gradient_compute:regularization}
As described in the previous section, the map $l \circ \gamma \circ \delta$ is differentiable. If the map $l$ is a function of a single element $(p,q)$ in a given persistence diagram $D_i$, the map $l \circ \gamma \circ \delta$ is locally a function of at most four elements of $X$. The elements in question are the two pairs of points maximum distance apart in the simplices whose introduction resulted in the appearance and disappearance of the topological feature in question. In this case, taking a single in the direction of the gradient of $l \circ \gamma \circ \delta$ will alter the positions of at most these four elements of $X$.

The above approach to minimizing a loss function changes topological features at the level of individual points. This does not agree with our prior that changes to topological features should be made at the level of larger entities consisting of sets of points. To overcome this challenge we propose a novel method for regularizing persistent homology gradient computation through the addition of a grouping term. This has the effect of helping to ensure gradients are defined with respect to larger entities and not individual points.

Let Set be the space of finite sets of points in $\mathbb{R}^n$. Let $X \in \text{Set}$ be the set of input points which we wish to optimize with respect to the loss function $l$. Let $X'$ be the initial value of $X$ before optimization and $\rho: \text{Set} \rightarrow \text{Set}$ be the corresponding bijection from $X'$ to $X$. Let $G$ be the set of all unordered pairs of elements in $X'$ and $k:[0,\infty] \rightarrow [0,1]$ be a given kernel. Broadly speaking, a kernel maps smaller values to a value approaching $1$ and larger values to a value approaching $0$. An example of a kernel is the \textit{uniform kernel} defined as follows where $s$ is a specified scale parameter:
\begin{equation}
	\label{eq:flat_kernel}
	k(x) = 
	\begin{cases}
		1 & \| x \| \leq s \\
		0 & \| x \| > s \\
	\end{cases}
\end{equation}
Given a specified kernel, we define the proposed regularization term $\tau: \text{Set} \rightarrow \mathbb{R}$ as follows:
\begin{equation}
	\label{eq:regularization_term}
	\tau(X) = \sum_{(a,b) \in G} k(\| a-b \|)(\| a-b \| - \| \rho(a)-\rho(b) \|)^2
\end{equation}
This term measures the discrepancy between pairwise distances in $X'$ and the corresponding pairwise distances in $X$. Minimizing this term helps to ensure the structure of local groups of points in $X'$ is preserved in $X$. That is, if a change is made to a single point this term will help ensure that a similar change is made to all points in a corresponding group where this group is a local neighbourhood defined by the kernel in question.

Let $\varrho: \text{Set} \rightarrow \mathbb{R}$ be a specified loss function which is a function of the persistent homology of the corresponding input. We define a regularized version of this loss $l: \text{Set} \rightarrow \mathbb{R}$, which integrates the regularization term in Equation \ref{eq:regularization_term}, as follows:
\begin{equation}
	\label{eq:regularization_loss}
	l(X) = \varrho(X) + \lambda \tau(X)
\end{equation}
In this equation $\lambda$ is a specified real valued weighting parameter. This regularized version of the loss will help ensure that changes to topological features will not be made at the level of individual points, but will instead be made at the level of larger entities consisting of sets of points. 

The result of applying the proposed regularization method to the sets of points in Figures~\ref{fig:two_cluster} and \ref{fig:circle} are displayed in Figures~\ref{fig:two_cluster_smooth} and \ref{fig:circle_smooth} respectively. In both cases a uniform kernel with a $s$ parameter value of $1.0$ and a $\lambda$ parameter value of $1.0$ was used. The loss function terms $\varrho$ in question equal the squared persistence of the element in $H_0$ corresponding to the merging of the two clusters and the squared persistence of the element in $H_1$ corresponding to the merging of the two ends of the horseshoe respectively. These functions are formally defined in the Appendix~\ref{appx:lossfunctions} section of this article. Optimization was performed using the Adam optimizer with a learning rate of $0.01$ \cite{kingma2015adam}.

From these figures, we see that the distance between the clusters and the distance between the ends of the horseshoe are reduced and this is achieved in a manner which agrees with our prior that changes to topological features are made at the level of larger entities. This contrasts with the results displayed in Figures~\ref{fig:two_cluster_zero_smooth} and \ref{fig:circle_zero_smooth} respectively where no regularization is applied and consequently changes to topological features are made at the level of individual points.

\section{Conclusions \& Future Work}
\label{sec:conclusions}
To the authors knowledge, this work presents the first attempt to perform regularization of persistent homology gradient computation. Given the increasing usefulness of integrating persistent homology and deep learning plus the need to perform such regularization, we believe this topic has the potential to develop into an active area of research in the field of applied topology.

\newpage
\appendix
\section{Loss Functions for Figure~\ref{fig:intro_shapes_single_component}}
\label{appx:lossfunctions}
In this section we formally define the loss function terms $\varrho$ in Equation \ref{eq:regularization_loss} applied to the datasets in Figures~\ref{fig:two_cluster} and \ref{fig:circle} to obtain the results in Figures~\ref{fig:two_cluster_zero_smooth}, \ref{fig:two_cluster_smooth}, \ref{fig:circle_zero_smooth} and \ref{fig:circle_smooth}.

Let $\mathcal{D}$ be the space of persistence diagrams. Let $\varrho_0: \mathcal{D} \rightarrow \mathbb{R}$ be a loss function defined as: 
\begin{equation}
	\label{eq:loss_0}
	\varrho_0(D) = \sum_{ \substack{ \lbrace (p,q) \in D : \\ q-p > 0.10, \\ q-p < \infty \rbrace } } (q-p)^2.
\end{equation}
This loss function was applied to the persistence diagram $D_0$ corresponding to the dataset in Figure~\ref{fig:two_cluster} to compute the results in Figures~\ref{fig:two_cluster_zero_smooth} and \ref{fig:two_cluster_smooth}.

Let $\varrho_0: \mathcal{D} \rightarrow \mathbb{R}$ be the loss function defined as:
\begin{equation}
	\label{eq:loss_1}
	\varrho_1(D) = \sum_{ \substack{ \lbrace (p,q) \in D : \\ q-p > 0.25 \rbrace } } (q-p)^2.
\end{equation}
This loss function was applied to the persistence diagram $D_1$ corresponding to the dataset in Figure~\ref{fig:circle} to compute the results in Figures~\ref{fig:circle_zero_smooth} and \ref{fig:circle_smooth}.

\end{document}